\documentclass[11pt]{article}
\usepackage{acl2016}
\usepackage{times}
\usepackage[hyphens]{url}
\usepackage{latexsym}
\usepackage{xspace}
\aclfinalcopy 
\pdfoutput=1

\newcommand{\bnhi}{\textit{bn-hi}\xspace}
\newcommand{\enhi}{\textit{en-hi}\xspace}
\newcommand{\tehi}{\textit{te-hi}\xspace}
\newcommand{\tahi}{\textit{ta-hi}\xspace}
\newcommand{\mrhi}{\textit{mr-hi}\xspace}

\title{Statistical Machine Translation for Indian Languages: Mission Hindi 2}

\author{Raj Nath Patel \\
	KBCS, CDAC Mumbai \\
	{\tt rajnathp@cdac.in} \\\And
	Prakash B. Pimpale \\
	KBCS, CDAC Mumbai \\
	{\tt prakash@cdac.in} \\}

\date{}

\begin{document}
\maketitle
\begin{abstract}
This paper presents Centre for Development of Advanced Computing Mumbai's (CDACM) submission to NLP Tools Contest on Statistical Machine Translation in Indian Languages (ILSMT) 2015 (collocated with ICON 2015). The aim of the contest was to collectively explore the effectiveness of Statistical Machine Translation (SMT) while translating within Indian languages and between English and Indian languages. In this paper, we report our work on all five language pairs, namely Bengali-Hindi (\bnhi), Marathi-Hindi (\mrhi), Tamil-Hindi (\tahi), Telugu-Hindi (\tehi), and English-Hindi (\enhi) for Health, Tourism and General domains. We have used suffix separation, compound splitting and preordering prior to SMT training and testing.
\end{abstract}


\section{Introduction}
\label{sec:intro}
In this paper, we present our SMT experiments from Bengali, Marathi, Tamil, Telugu and English to Hindi. From the set of languages involved, Bengali, Hindi and Marathi belong to the Indo-Aryan family and Tamil and Telugu are from Dravidian language family. All languages except English, have the same flexibility towards word order, canonically following
the Subject-Object-Verb (SOV) structure.

With reference to morphology Bengali, Marathi, Tamil, and Telugu are more agglutinative compared to Hindi. It is known that SMT produces unknown words resulting in bad translation quality if morphological divergence between source and target languages is high. ~\newcite{koehn:2003:CS}, ~\newcite{popovic:2004:word-stem}, nd~\newcite{popovic:2006:CW} have demonstrated ways to handle this issue with morphological segmentation of words in the source sentences before training the SMT system. To handle this morphological difference we have used suffix separation and compound word splitting developed by~\newcite{pimpale:2014} and~\newcite{patel:2014}.

For English-Hindi SMT, we achieve better alignment using preordering~\cite{patel:2013} and stem as an alignment factor~\cite{koehn:2007:factored}.

All machine translation (MT) systems suffer from Out of Vocabulary (OOV) words. These OOV words are mostly named entities, technical terms and foreign words which can be translated using transliteration system. We used ~\newcite{durrani:2014}, which is a fully unsupervised approach for developing a transliteration system using parallel corpus meant for the SMT training. 

The rest of the paper is organized as follows. In section~\ref{sec:settings}, we discuss dataset and experimental setup. Section~\ref{sec:results} discusses experiments and results. Submitted systems to the shared task are described in section~\ref{sec:submit} followed by the conclusion and future work in section~\ref{sec:conclution}.

\section{Data-set and Experimental Setup}
\label{sec:settings}
In the following subsections, we describe Training and Testing corpus followed by pre-processing and SMT system setup for experiments.

\subsection{Corpus for SMT Training and Testing}
\label{subsec:data}
We have used corpus shared by ILSMT detailed in Table~\ref{tab:data} for the experiments. Testing for the experiments was done using Test1 which is the development set. The submitted systems were evaluated against Test2 corpus, by the organizers. For unconstrained systems, additional data~\cite{bojar:2014:monohindi,khapra:2010} has been used for language modeling.

\begin{table}[!hbt]
	\centering
	\begin{tabular}{l|c|c|c}
			& health & tourism & general \\ \hline
	training (TM) & 24K & 24K & 48K \\
	training (LM) & 24K & 24K & 48K \\
	test1 & 500 & 500 & 1000 \\
	test2 & 500 & 400 & 1000 \\
	\end{tabular}
	\caption{Corpus distribution; TM: Translation Model, LM: Language Model}
	\label{tab:data}
\end{table}

\subsection{SMT System Set Up}
\label{subsec:setup}
The baseline system was setup by using the phrase-based model~\cite{och:2003,brown:1990,marcu:2002,koehn:2003,koehn:2007} and~\newcite{koehn:2007:factored} was used for factored model. The language model was trained using KenLM~\cite{heafield:2011} toolkit with modified Kneser-Ney smoothing~\cite{chen:1996}. For factored SMT training source and target side stem were used as an alignment factor. Stemming was done using a lightweight stemmer for Hindi~\cite{ramanathan:2003:stem}. For English, we used porter stemmer~\cite{minnen:2001}.

\subsection{Evaluation Metrics}
\label{subsec:eval}
The different experimental systems were compared using, BLEU~\cite{papineni:2002}, NIST~\cite{doddington:2002}, translation error rate (TER)~\cite{snover:2006:ter}. For an MT system is to be better, higher BLEU and NIST scores with lower TER are desired.

\subsection{Pre-Processing (PP)}
\label{subsec:pp}
To tackle the morphological divergence between the source and target languages for the purpose of a better SMT system, we preprocessed the source (Bengali, Marathi, Tamil, and Telugu) for suffix separation and compound word splitting ~\cite{pimpale:2014,patel:2014} prior to training and testing. To handle the structural divergence (\enhi), we used source side reordering~\cite{patel:2013}.

\subsection{Transliteration (TR)}
\label{subsec:tr}
We developed transliteration systems using ~\newcite{durrani:2014} inbuilt with the Moses tool. We used n-best transliteration output for OOV words. These candidates were then plugged in and re-scored with the language model to get the best translation for the given source sentence.

\subsection{Language Modelling}
\label{subsec:lm}
We trained 5-gram LM using KenLM with modified Kneser-Ney smoothing. LM size was quite large (ARPA-88.7GB, binary-27.4GB), even after binarization. Training LM with the huge amount of monolingual data like ~\newcite{bojar:2014:monohindi} (approx. 10GB ) required good computing resources (60GB RAM and 200GB storage space in the working dir). Also, ~\newcite{bojar:2014:monohindi} corpus contains unwanted symbols ($<$s$>$) for KenLM which we need to remove prior training.


\section{Experiments and Results}
\label{sec:results}
Table~\ref{tab:results:cons} shows evaluation scores for various systems tried under constrained submission, that is, systems trained only on the shared data. We can see the use of the preprocessing and transliteration contributed to the improvement of around 1-5 bleu points across the language pairs. Detailed evaluation scores for unconstrained systems, that is, systems using language model built on external data, are in Table~\ref{tab:results:uncons}. Significant improvement can be observed from the table when additional data was used for language modeling.

\begin{table}[!hbt]
	\centering
	\begin{tabular}{l|c|c|c|c}
		 &  & BLEU &  NIST & TER \\ \hline
	\bnhi & S1 & 30.18 & 6.888 & 48.43 \\
		& S2 & 31.23 & 6.943 & 47.27 \\
		& S3 & 32.22 & 7.092 & 46.46 \\ \hline
	\enhi & S1 & 18.76 & 5.862 & 64.05 \\
		& S2 & 23.15 & 6.037 & 59.89 \\
		& S3 & 23.44 & 6.092 & 59.58 \\ \hline
	\mrhi & S1 & 35.74 & 7.511 & 42.75 \\
		& S2 & 40.01 & 7.805 & 39.95 \\
		& S3 & 40.21 & 7.867 & 39.62 \\ \hline
	\tahi & S1 & 16.64 & 4.742 & 64.68 \\
		& S2 & 20.17 & 5.303 & 62.65 \\
		& S3 & 20.58 & 5.391 & 62.12 \\ \hline
	\tehi & S1 & 24.88 & 6.219 & 52.18 \\
		& S2 & 28.94 & 6.532 & 49.97 \\
		& S3 & 29.86 & 6.679 & 49.19
	\end{tabular}
	\caption{Scores for different systems (CONSTRAINED); S1:BL; S2:BL+PP; S3:BL+PP+TR; BL:Baseline; TR:Transliteration}
	\label{tab:results:cons}
\end{table}

\begin{table}[!hbt]
	\centering
	\begin{tabular}{l|c|c|c|c}
		&  & BLEU &  NIST & TER \\ \hline
		\bnhi & S1 & 30.18 & 6.888 & 48.43 \\
		& S2' & 31.58 & 7.080 & 47.03 \\
		& S3' & 33.77 & 7.195 & 45.52 \\ \hline
		\enhi & S1 & 18.76 & 5.862 & 64.05 \\
		& S2' & 19.96 & 5.922 & 63.03 \\
		& S3' & 24.00 & 6.121 & 58.99 \\ \hline
		\mrhi & S1 & 35.74 & 7.511 & 42.75 \\
		& S2' & 36.80 & 7.617 & 42.35 \\
		& S3' & 41.20 & 7.935 & 39.25 \\ \hline
		\tahi & S1 & 16.64 & 4.742 & 64.68 \\
		& S2' & 16.38 & 4.677 & 64.45 \\
		& S3' & 20.38 & 5.340 & 62.25 \\ \hline
		\tehi & S1 & 24.88 & 6.219 & 52.18 \\
		& S2' & 25.47 & 6.251 & 51.99 \\
		& S3' & 29.72 & 6.669 & 49.26 \\
	\end{tabular}
	\caption{Scores for different systems (UNCONSTRAINED); S1:BL; S2':BL+ELM; S2':BL+PP+ELM; BL:Baseline; ELM:Extended LM (Additional data used for LM training)}
	\label{tab:results:uncons}
\end{table}


\section{Submission to the Shared Task}
\label{sec:submit}
We submitted two different results for the contest, namely constrained and unconstrained. The constrained systems were trained only on the data shared by the organizers. For unconstrained systems, we used additional monolingual data which include~\newcite{bojar:2014:monohindi} and~\newcite{khapra:2010}, for language modeling.

\section{Conclusion and Feature Work}
\label{sec:conclution}	
In this paper, we presented systems for translation from Bengali, English, Marathi, Tamil and Telugu to Hindi. These SMT systems with the use of source side suffix separation, compound splitting, preordering and bigger language model shows significantly higher accuracy over the baseline. Adding more complex features for factored models and formulating preprocessing with better way could be the next step.

\bibliography{acl2016}
\bibliographystyle{acl2016}

\end{document}